\documentclass{sig-alternate-05-2015}
\usepackage[english]{babel}
\usepackage{etoolbox}
\usepackage[font={small}]{caption}
\usepackage{amsmath}
\usepackage{enumerate}
\usepackage{bm}
\usepackage[numbers]{natbib}
\usepackage{url}
\usepackage{amssymb} 
\usepackage{amsmath}
\usepackage{graphicx}
\usepackage{caption}
\usepackage{enumitem}
\usepackage{subcaption}
\usepackage[table]{xcolor}
\definecolor{lightgray}{gray}{0.85}
\usepackage{algorithmicx}
\usepackage{algorithm}
\usepackage{algpseudocode}
\usepackage[font=small]{caption} 

\setlist[itemize]{noitemsep,topsep=2pt}
\usepackage{tabularx}
\usepackage{enumitem} 
\usepackage{lipsum}

\let\OLDthebibliography\thebibliography
\renewcommand\thebibliography[1]{
  \OLDthebibliography{#1}
  \setlength{\parskip}{2pt}
  \setlength{\itemsep}{2pt plus 0.3ex}
}
\usepackage{multirow}
\usepackage{todonotes}

\makeatletter
\def\@copyrightspace{
    \@float{copyrightbox}[b]
    \begin{center}
        \setlength{\unitlength}{0.2pc}
            \begin{picture}(100,6) 
                \put(0,-6){\crnotice{\@toappear}}
            \end{picture}
        \end{center}
    \end@float}
\makeatother

\begin{document}





\conferenceinfo{Accepted at 4th International Workshop on SWDM, co-located with CIKM}{October, 2016, Indianapolis, USA}


\title{Semi-supervised Discovery of Informative Tweets During the Emerging Disasters}
%
%
%
%
%
%

\numberofauthors{2} 
%
\author{
%
%
\alignauthor
Shanshan Zhang\\
       \affaddr{Temple University}\\
       \affaddr{1805 N. Broad Street}\\
       \affaddr{Philadelphia, PA}\\
       \email{zhang.shanshan@temple.edu}
\alignauthor
Slobodan Vucetic\\
	   \affaddr{Temple University}\\
       \affaddr{1805 N. Broad Street}\\
       \affaddr{Philadelphia, PA}\\
       \email{vucetic@temple.edu}
}

\maketitle
\begin{abstract}
The first objective towards the effective use of microblogging services such as Twitter for situational awareness during the emerging disasters is discovery of the disaster-related postings. Given the wide range of possible disasters, using a pre-selected set of disaster-related keywords for the discovery is suboptimal. An alternative that we focus on in this work is to train a classifier using a small set of labeled postings that are becoming available as a disaster is emerging. Our hypothesis is that utilizing large quantities of historical microblogs could improve the quality of classification, as compared to training a classifier only on the labeled data. We propose to use unlabeled microblogs to cluster words into a limited number of clusters and use the word clusters as features for classification. To evaluate the proposed semi-supervised approach, we used Twitter data from 6 different disasters. Our results indicate that when the number of labeled tweets is 100 or less, the proposed approach is superior to the standard classification based on the bag or words feature representation. Our results also reveal that the choice of the unlabeled corpus, the choice of word clustering algorithm, and the choice of hyperparameters can have a significant impact on the classification accuracy. 
\end{abstract}

%
%

\ccsdesc[500]{Information systems~Data analytics}

%
%

%
%
\printccsdesc


\keywords{semi-supervised learning; Twitter; crisis management}

\section{Introduction}
During the past decade, there has been plentiful evidence that social media can be very useful in improving the situational awareness during emergencies and disasters \cite{IM15}, and several research groups even proposed specific systems to extract, process, and summarize social media data for such purpose \cite{AF12, AZ14, IM14}. The prototypical situational awareness system relying on social media, such as the popular microblogging service Twitter, consists of a data collection, a data processing, and a visual interface component. The first critical challenge in making those systems work is selection of the disaster-related posts from a massive stream of posts pouring from services such as Twitter. The objective of this work is to propose a new approach for selection of informative posts, with a particular focus on the beginning stages of a disaster, when very little is known about the emergency and the ways in which people report about and react to it.

There are two main approaches for selection of the informative posts. One is the information retrieval approach, which is based on developing a lexicon of disaster-related unigrams or bigrams based on a study of posts from previous disasters and emergencies \cite{OA14, TI15}. While there are terms that are commonly used over different disaster types (e.g. ``victims'', ``prayers'', ``help''), each disaster is different and might require its own specialized lexicon. As such, pre-built lexicons cannot be expected to have very large sensitivity and specificity during the emerging disasters. An alternative supervised learning approach is based on training a classifier that recognizes informative posts \cite{AF12,AZ14, IM14}. To train the classifier, there is a need to have access to labeled posts. If a sufficient number of labeled posts are available, the supervised approach is superior to the keyword-based approach. However, during the first few minutes or hours from the emergence of a disaster, the number of labeled posts might be too small to result in accurate classification. 

The main issue when training from small labeled data is avoiding to build overly complex models that cannot generalize well. A typical remedy is to keep classification models simple. That is why logistic regression is a very popular model choice in the low-sample learning scenarios. A closely related issue is that the dimensionality of data is often too large and it complicates the issue of learning from small labeled data. For example, in case of classifying text data such as microblogs, a popular data representation approach is the bag of words representation, in which each word from a dictionary is treated as a feature. A standard remedy is to reduce dimensionality by feature selection and use some form of model regularization. However, when the number of labels is very small (less than a hundred), selection of the most informative words to be used as features can become unreliable. In particular, for all but the most frequent words, there might be too little evidence to estimate how useful they are with any degree of statistical certainty. 

In this work, we propose to cluster similar words in groups and treat each cluster as a feature. By treating word clusters as features, the features would become less sparse and the estimate of their usefulness for classification would be more confident. In addition, word clusters would allow utilizing words that occur rarely or do not even occur in the labeled corpus. To create word clusters, we rely on unlabeled historical posts. One of the open questions in this study was what kind of unlabeled corpus is the most appropriate for this application. In the following text, we describe the details of our approach and provide an experimental evaluation of the approach on Twitter data from several disasters.

\section{Methodology}
\subsection{Problem Setup}
Let us suppose we are given a corpus of $n$ labeled documents $\textbf{D}_{train}$ = $\{(d_1, y_1), (d_2, y_2), ..., (d_n, y_n)\}$ and each document $d_i$ is converted into a feature vector $\boldsymbol{x_i}\in \mathbb{R}^m$. For example, we could use the bag of words representation, where given a vocabulary $\textbf{V}$ the feature vector $\boldsymbol{x_i}$ is a binary vector of length $|\textbf{V}|$, whose $j$-th element indicates whether the $j$-th word from the vocabulary is present in the document $d_i$. By a slight extension, \textit{n-}grams, part-of-speech tags, named entities, or word clusters could also be used for feature representation of documents. The label $y_i$ of document $d_i$ is a binary variable indicating whether it is disaster-related or not, where $y_i=1$ indicates that the document is disaster related and $y_i=0$ that it is not. Let us also assume that we are given a corpus of unlabeled documents $\textbf{D}_{unl} = \{d_1, d_2, ..., d_N\}$ containing $N$ documents that are available at the training time. The objective is to train a model $f(x)$ whose output can be used to classify a document. We will assume that the output is real-valued and that larger values indicate a stronger likelihood of being disaster-related. For classification, we need to select a particular threshold $\theta$ and classify all documents with outputs above the threshold as disaster-related and those below the threshold as unrelated.

\subsection{Feature Selection Filters}

The objective of feature selection is to reduce dimensionality of the feature vector. Feature selection filters are the simplest class of such algorithms that attempt to select only a subset of the most informative features and remove the rest. They do it so by calculating the score of each feature that measures how strong its correlation is with the classification label. Only the $K$ features with the largest score are retained. There are many known scoring functions \cite{RP04}, among which the $\chi^2$ statistics and Pointwise Mutual Information (PMI) are very popular for text classification. In this paper, we will use the PMI score, which is defined in the following way:
\begin{equation}
score(t) = PMI(t, pos) - PMI(t, neg) = \log_2\frac{p(t|pos)}{p(t|neg)}
\end{equation}
where $p(t|pos)$ is the probability of feature $t$ appearing in positive class, and $p(t|neg)$ is the probability of feature $t$ appearing in negative class.

\subsection{Word Clusters as Features}
The idea of document representation with word clusters is to group similar words and treat each cluster as a feature. In particular, given a vocabulary $\textbf{V}$ the objective is to assign them to one of $K$ clusters. Given those clusters, each document $d_i$ is converted into a binary feature vector $\boldsymbol{x_i}$ of length $K$ whose $j$-th element $x_{ij}$ equals one if any of the words from the $j$-th cluster are present in the document and zero if not. 

To create the clusters, we rely on the corpus of unlabeled documents $\textbf{D}_{unl} = \{d_1, d_2, ..., d_N\}$. Given the corpus, we will evaluate two ways of generating word clusters, as explained next. 

\textbf{Brown clustering (BC)}\cite{BPF92}. This is a traditional algorithm for word clustering. It is a hierarchical clustering algorithm that assigns a single cluster to each word and proceeds by merging the two clusters that result in the smallest reduction of global mutual information. The output of Brown clustering is a dendrogram of words, which means we can cut at any level in the hierarchy to create word clusters.

\textbf{Clustering based on the skip-gram representation of words}. The idea of neural language models, of which the skip-gram model \cite{MT13} is a representative, is to learn low-dimensional vector representation of words. The skip-gram model defines the probability of word $w_N$ being in the neighbourhood of a target word $w_T$ in text by assuming that the words in the neighborhood are conditionally independent given the target word. It models probability of neighbour $w_N$ given the target word $w_T$ as a softmax function
\begin{equation}
p(w_{N} | w_T) = \frac{exp({v^{'}_{w_N}}^T v_{w_T})}{\sum_{w=1}^{|\textbf{V}|}exp({v^{'}_{w}}^T v_{w_T})}
\end{equation}
where $v_{w} \in \mathbb{R}^{p}$ and $v^{'}_{w} \in \mathbb{R}^{p}$ are the ``input'' and ``output'' $p$-dimensional vector representations of word $w$. The input and output representations of every word are obtained by maximizing the log-likelihood of the model over a corpus of documents. In this paper, we use word2vec \cite{MT13}, which is a stochastic gradient algorithm that is commonly used to maximize this objective function. The resulting word representations place words occurring in similar contexts near each other. 

Given the output word representations from the word2vec algorithm, we use the traditional k-means clustering algorithm to group all words form the vocabulary into $K$ clusters.

\section{Experiments and Evaluation}

The objective of our experimental evaluation is to obtain an insight into the performance of the proposed semi-supervised approach, which uses an unlabeled document corpus to create word clusters and trains a classifier on a small labeled document corpus where the word clusters are used as features.

\subsection{Data Set}

For our experiments, we used CrisisLexT6 data described in \cite{OA14}. It is a collection of tweets related to 6 natural disasters that occurred between October 2012 and July 2013: Sandy Hurricane (\textbf{SH}), Boston Bombings (\textbf{BB}), Oklahoma Tornado (\textbf{OT}), West Taxas Explosion (\textbf{WE}),Alberta Floods (\textbf{AF}), Queensland Floods (\textbf{QF}). 

For each disaster, the same number of tweets was selected using two different methods. One method relied on a predefined set of keywords appropriate for a particular disaster. Another method relies on selecting geocoded tweets originating from the affected areas during the onset of a disaster. The collected tweets were labeled as disaster or not disaster related by 100 workers from Crowdflower. Out of the 60,000 Twitter IDs of the labeled tweets provided by the authors of \cite{OA14}, we were only able to download 70\% of them and we summarize their basic statistics in Table \ref{tab:datasummary}.

\begin{table}[!h]
	\begin{center}
	\tiny
	\caption{Basic data statistics of each disaster.}
	\label{tab:datasummary}
		\begin{tabular}{l<{\hspace{-3pt}}|c<{\hspace{-3pt}}|c<{\hspace{-3pt}}||c<{\hspace{-3pt}}|c<{\hspace{-3pt}}}
			\hline
			\hline
			\textbf{Disaster} & \textbf{\# Location-} & \multirow{2}{*}{\textbf{\# Positive}} & \textbf{\# Keyword-} & \multirow{2}{*}{\textbf{\# Positive}}\\
			\textbf{Type} & \textbf{based sample} & & \textbf{based sample} & \\\hline
			\textbf{Sandy} & \multirow{2}{*}{\small 3505}& \multirow{2}{*}{\small 1000} &  \multirow{2}{*}{\small 3267} & \multirow{2}{*}{\small 3058} \\
			\textbf{Hurricane} & & & & \\\hline
			\textbf{Boston} & \multirow{2}{*}{\small 3604}& \multirow{2}{*}{\small 610} &  \multirow{2}{*}{\small 3651} & \multirow{2}{*}{\small 3451}\\
			\textbf{Bombings} & & & & \\\hline
			\textbf{Oklahoma} & \multirow{2}{*}{\small 4088}& \multirow{2}{*}{\small 423} &  \multirow{2}{*}{\small 3897} & \multirow{2}{*}{\small 3362}\\
			\textbf{Tornado} & & & & \\\hline
			\textbf{West Taxas} & \multirow{2}{*}{\small 4535}& \multirow{2}{*}{\small 305} &  \multirow{2}{*}{\small 4245} & \multirow{2}{*}{\small 4168} \\
			\textbf{Explosion} & & & & \\\hline
			\textbf{Alberta} & \multirow{2}{*}{\small 3712}& \multirow{2}{*}{\small 367} &  \multirow{2}{*}{\small 4384} & \multirow{2}{*}{\small 4044}\\
			\textbf{Floods} & & & & \\\hline
			\textbf{Queensland} & \multirow{2}{*}{\small 3851}& \multirow{2}{*}{\small 361} &  \multirow{2}{*}{\small 4302} & \multirow{2}{*}{\small 4051}\\
			\textbf{Floods} & & & & \\\hline
		\end{tabular}
	\end{center}
\end{table}

From Table \ref{tab:datasummary}, we observe a highly skewed distribution of classes depending on the selection criteria. In particular, the keyword-based sample contains a high fraction of positive (disaster-related) labels, while the location-based sample has a high fraction of negative labels. Unlike the previous work that merged tweets from those two samples and split them randomly into training and test sets for evaluation \cite{IM14, IM16, OA14}, in this work we used only the tweets obtained by the location search. We discarded the portion of tweets obtained by the keyword-based selection. Our reasoning was that as long as the keywords are well defined by authorities, virtually all tweets sampled by the keyword search can be assumed as positives. Identifying disaster-related Tweets within the location-based sample is a more challenging classification problem, which can be particularly important during the initial minutes or hours of an emerging disaster. 

\subsection{Experiment Design}

In our experiments we wanted to compare the proposed approach with the standard supervised learning approach that uses feature selection on the bag of words and trains a logistic regression classifier. We also wanted to explore impact of a range of choices on the accuracy, including the choice of the word clustering algorithm, the choice of an unlabeled document corpus, the size of the labeled corpus, the number of features, and the choice of hyperparameters.

\textit{Preprocessing}. All tweets were tokenized using the tool from twitter\_nlp\footnote{https://github.com/aritter/twitter\_nlp}. Stop-words, URLs, and user mentions (\textit{@username}), tokens too short (2 characters or less), tokens too long (16 characters or more) were removed from the tweets. Given the tweets preprocessed in this way, we divided the location-based tweets of each disaster randomly into two subsets: 70\% as training set and 30\% as the test set. The test set was not used in any stage of classifier building and was used solely to calculate the accuracy. 

\textit{Classification Algorithm}. In all experiments we relied on Logistic Regression (\textbf{LR}) with $L_2$ regularization. The regularization hyperparameter was obtained using leave-one-out cross-validation on the training data. 

\textit{Choice of training size}. We varied the training size in the range $t=[20, 50, 100, 200, 500, 1000]$. This allowed to achieve an insight into the impact of the training size on accuracy and its relative impact on different algorithms. The smallest training sizes of $20$ and $50$ are representative of scenarios with extremely limited labeled data sets that might be expected shortly after the outset of a disaster, while the larger training sizes are representatives of labeled data sets that can become available with a more significant delay.

\textit{Choice of features}. We compared two clustering algorithms described in the Methodology, the Brown Clustering (\textbf{BC}) and the clustering based on word2vec (\textbf{W2V}). We denote feature selection based on the traditional bag-of-words as \textbf{BOW}. We varied the number of clusters in the range of $K=[50, 100, 200, 500, 1000, 2000]$. To be comparable, for \textbf{BOW} we used feature selection based on PMI score to train Logistic Regression based on the top $K$ scoring words.

\textit{Choice of the unlabeled corpus for word clustering}. We explored several types of unlabeled Twitter sets for word clustering using \textbf{BC} and \textbf{W2V} approaches:
 
\begin{itemize}
\item $\textbf{D}_{50K}$. For any disaster, the unlabeled corpus contains all available CrisisLexT6 tweets from the other 5 disasters.
\item $\textbf{D}_{50Kp}$. For any disaster, the unlabeled corpus contains all positively labeled CrisisLexT6 tweets from the other 5 disasters.
\item $\textbf{D}_{3K}$. For any disaster the corpus contains all available keyword-based tweets from that disaster.
\end{itemize}

In addition, we downloaded a pre-trained \textit{word2vec} matrix $\textbf{W}$ with word vector dimension $p = 400$ and vocabulary size $|\textbf{V}| \sim 3,000,000$ that was trained on 400 million Tweets. 

By combining different choices for feature construction, we obtained 8 different algorithms that we evaluated experimentally: bag of words (\textbf{BOW}), bag of Brown clusters ($\textbf{BC}_{50K}$, $\textbf{BC}_{50Kp}$, $\textbf{BC}_{3K}$), and bag of word2vec clusters ($\textbf{W2V}_{50K}$, $\textbf{W2V}_{50Kp}$, $\textbf{W2V}_{3K}$, $\textbf{W2V}_{400M}$)

\textit{Other experimental settings}. We run \textit{word2vec} with negative sampling rate $10^{-3}$, context window size 100, and word vector dimension 20 (due to a relatively small unlabeled corpus). For each choice of the feature construction algorithms, each training set size, and each number of features $K$, we run Logistic Regression 10 times on randomly sampled subsets from the training set. The average Area Under the ROC Curve (AUC) of the 10 random sampling experiments is reported.

\subsection{Results}

\begin{figure*}[!htb]
	\centering
	\begin{subfigure}[t]{.45\textwidth}
		\includegraphics[scale = 0.45]{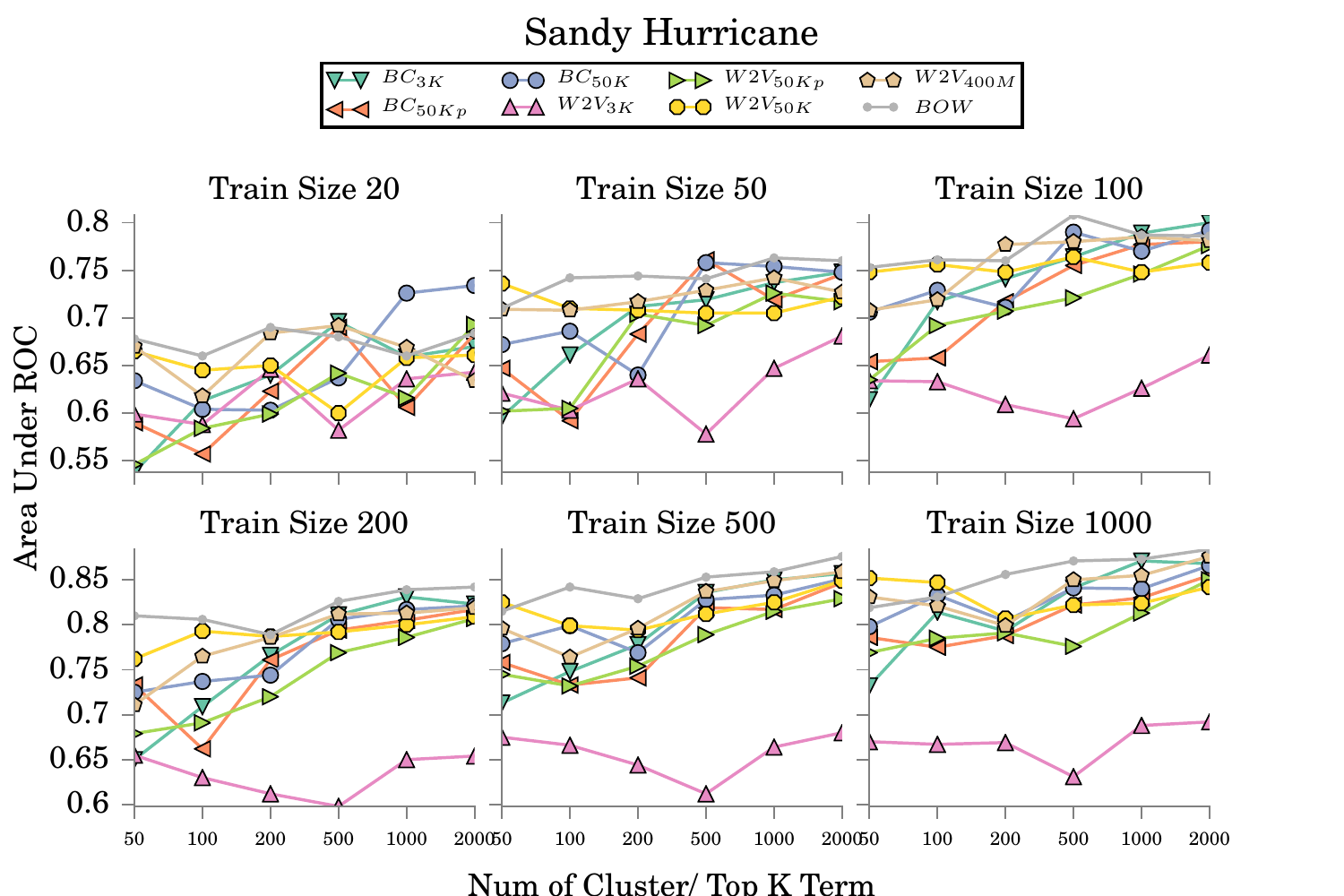}
	\end{subfigure}
	\begin{subfigure}[t]{.45\textwidth}
		\includegraphics[scale = 0.45]{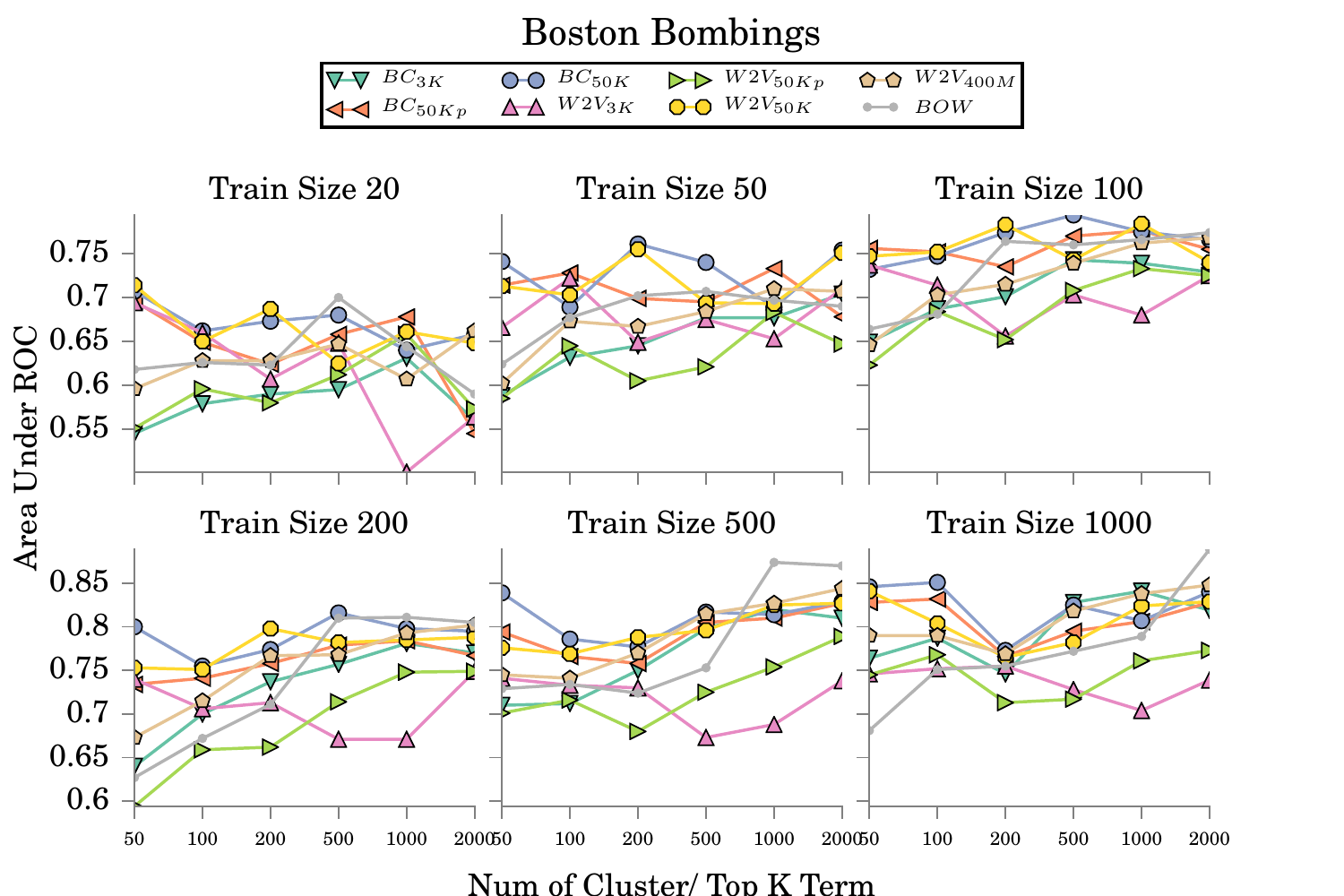}
	\end{subfigure}
	
	\begin{subfigure}[t]{.45\textwidth}
		\includegraphics[scale = 0.45]{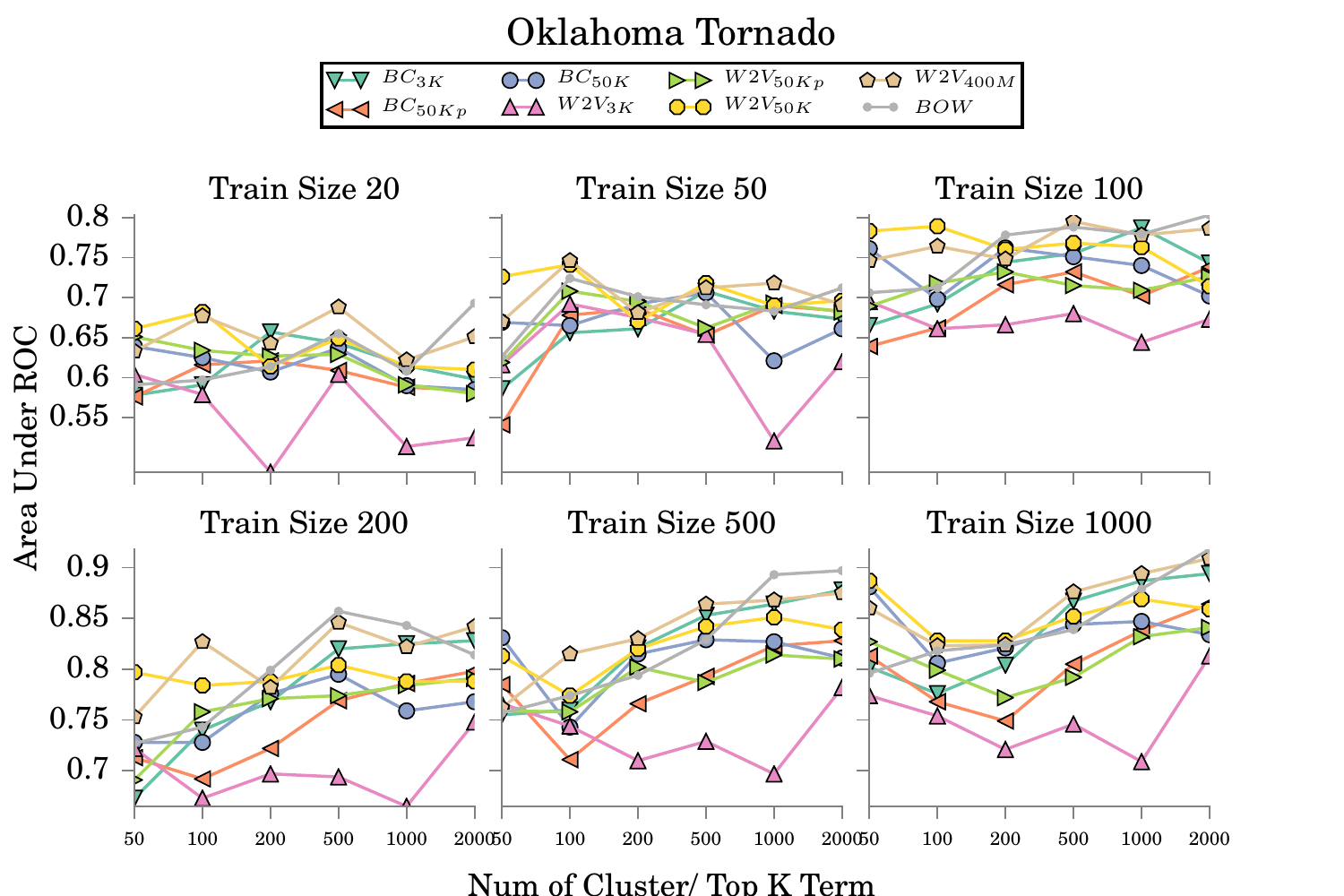}
	\end{subfigure}	
	\begin{subfigure}[t]{.45\textwidth}
		\includegraphics[scale = 0.45]{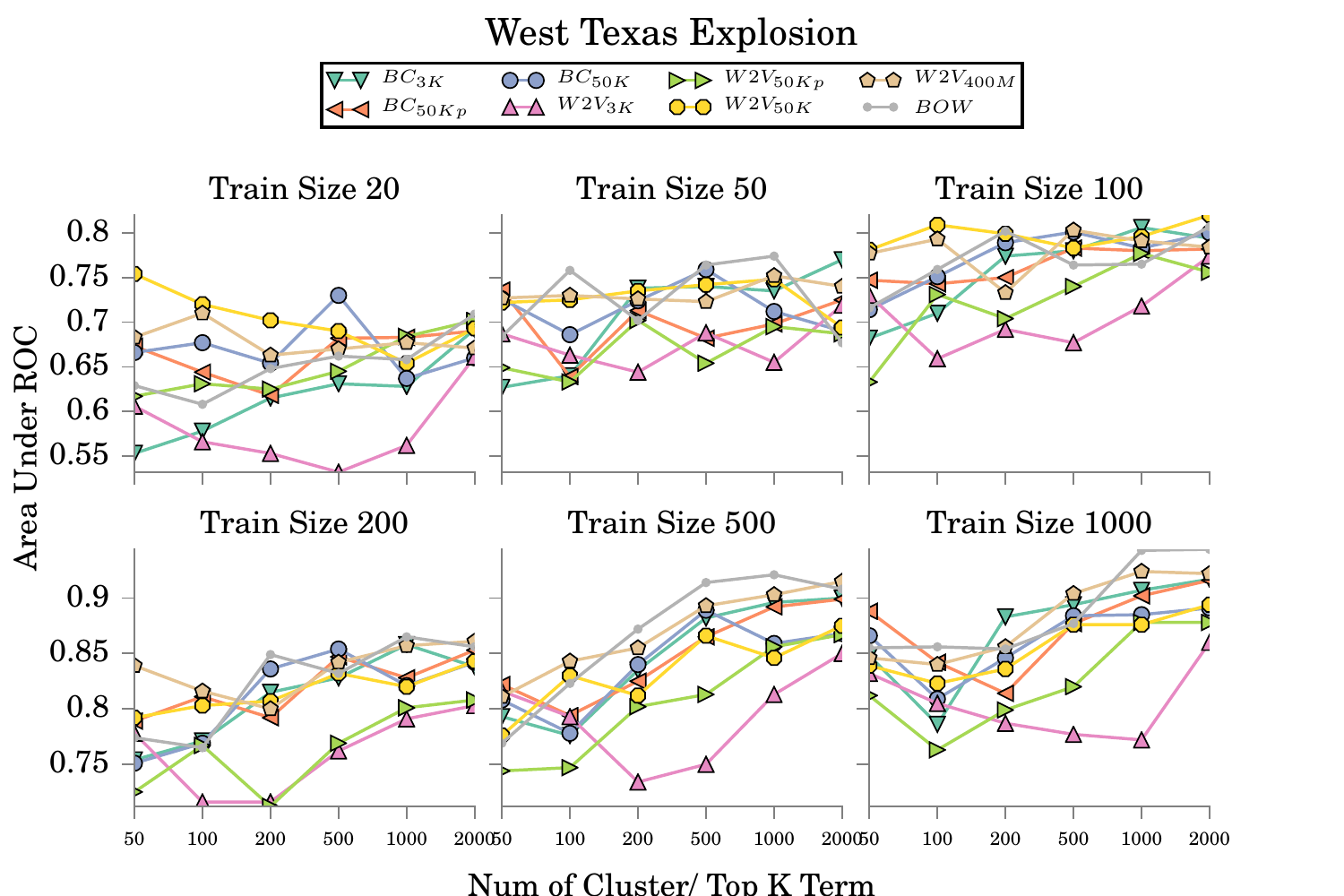}
	\end{subfigure}	

	\begin{subfigure}[t]{.45\textwidth}
		\includegraphics[scale = 0.45]{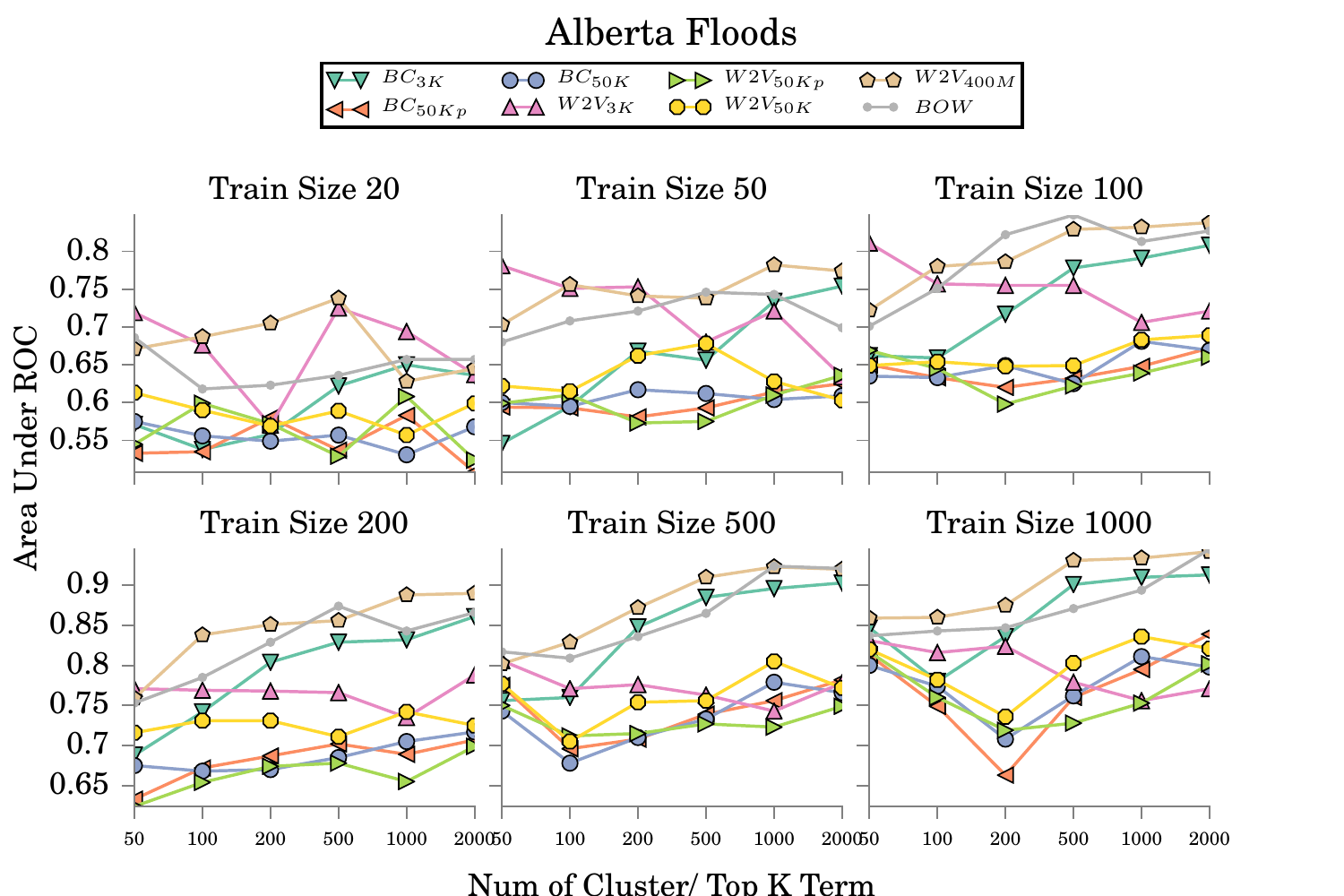}
	\end{subfigure}
	\begin{subfigure}[t]{.45\textwidth}
		\includegraphics[scale = 0.45]{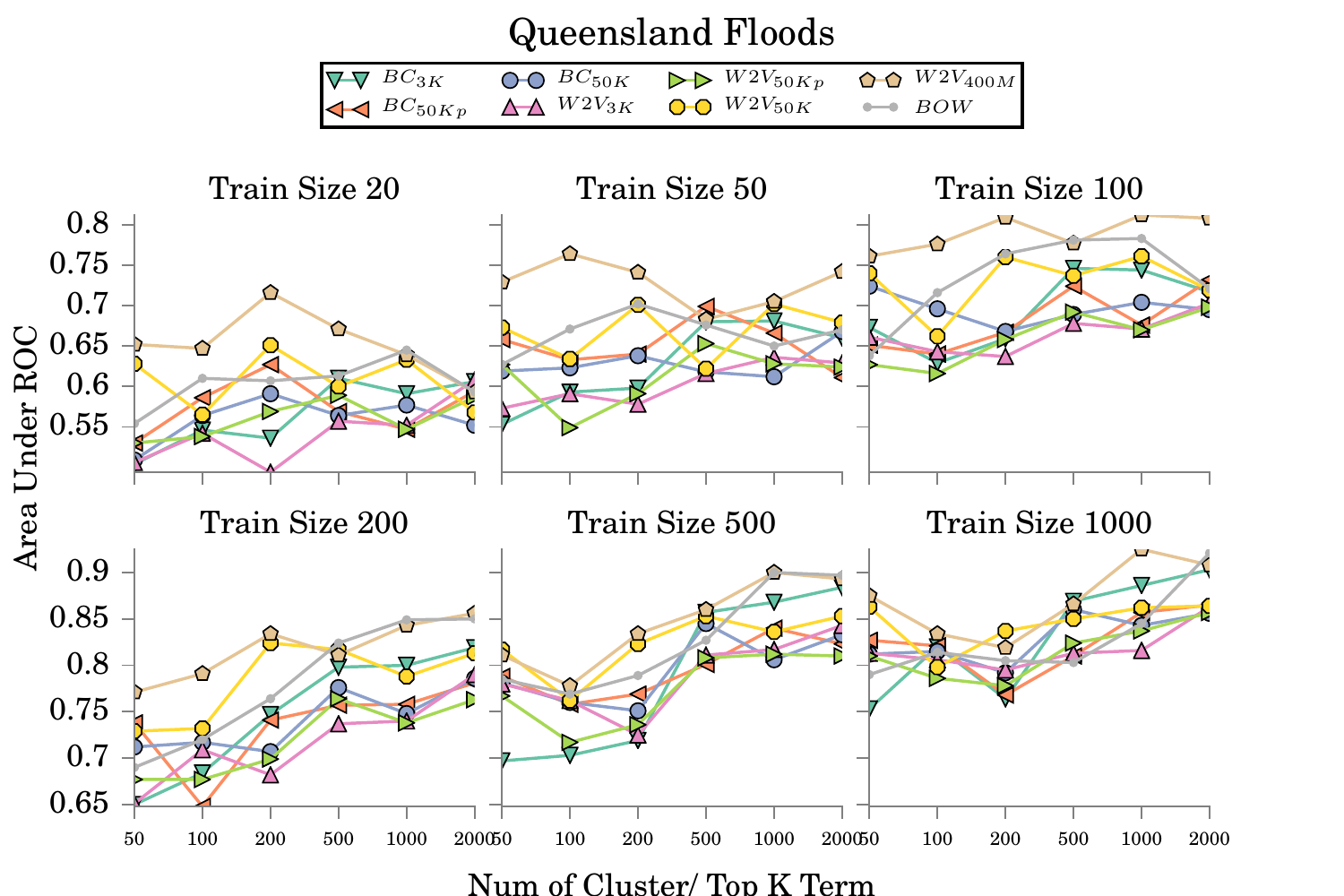}
	\end{subfigure}
    \caption{Detailed performance comparison for 6 disasters. }
    \label{fig:performance}
\end{figure*}

Our results provide insights into several interesting questions related to recognizing disaster-related tweets during the emerging disasters. Figure \ref{fig:performance} is the most comprehensive view into our results. For each disaster and each training size, we are showing 8 curves, representing 8 different feature construction algorithms, as a function of the number of features. However, to answer our posed questions, we summarized those results with 4 tables, as will be described in the following.

\textbf{Are word clusters better than \textbf{BOW}}? In Table \ref{tab:ssvsbow}, we compare the accuracies of the best word clustering and the best \textbf{BOW} classifier for each of the 6 disasters and for each training data size. Notice that for \textbf{BOW} approach at a specific training size, we report the best accuracy among different choices of $K$, which means we are treating the $K$ as a hyperparameter during training. For the 7 semi-supervised approaches, we report the best accuracy from that group. When the training data size is $20$, $50$, and $100$, the word clustering approach is superior to \textbf{BOW} on all but the Oklahoma Tornado (\textbf{OT}) data set. This result strongly indicates that an unlabeled Twitter corpus could provide a very useful information when there is a severe deficit of labeled tweets for an emerging disaster. As the number of labeled tweets is exceeding $100$, the benefit from the unlabeled corpus fades, to the extent that \textbf{BOW} is more accurate than the word clusters on all 6 disasters when the training data size reaches $1,000$. This is an expected result, since such a large training data set is sufficient to identify the most informative disaster-related words. 

\begin{table}[!htbp]
	\begin{center}
	\small
	\addtolength{\tabcolsep}{-4.0pt}
	\caption{Summary results to compare Semi-supervised (\textbf{SS}) and \textbf{BOW} algorithms.}
	\label{tab:ssvsbow}
		\begin{tabular}{l|c|c||c|c||c|c}
		\hline
		\hline
		\multirow{2}{*}{\textbf{Train}} & \multicolumn{2}{c||}{\textbf{SH}} & \multicolumn{2}{c||}{\textbf{BB}} & \multicolumn{2}{c}{\textbf{OT}} \\\cline{2-7}
			 & \textbf{$\textbf{SS}^*$} & \textbf{$\textbf{BOW}^*$} & \textbf{$\textbf{SS}^*$} & \textbf{$\textbf{BOW}^*$} & \textbf{$\textbf{SS}^*$} & \textbf{$\textbf{BOW}^*$} \\\hline
			\textbf{20} & $\boldsymbol{0.734}$ & $0.69$ & $\boldsymbol{0.714}$ & $0.7$ & $0.688$ & $0.693$ \\\hline
			\textbf{50} & $\boldsymbol{0.76}$ & $\boldsymbol{0.76}$ & $\boldsymbol{0.761}$ & $0.707$ & $0.746$ & $0.746$ \\\hline
			\textbf{100} & $\boldsymbol{0.8}$ & $\boldsymbol{0.8}$ & $\boldsymbol{0.794}$ & $0.774$ & $0.795$ & $0.803$ \\\hline
			\textbf{200} & $0.838$ & $0.844$ & $\boldsymbol{0.816}$ & $0.811$ & $0.846$ & $0.857$ \\\hline
			\textbf{500} & $0.865$ & $0.878$ & $0.844$ & $0.874$ & $0.887$ & $0.897$ \\\hline
			\textbf{1000} & $0.883$ & $0.892$ & $0.868$ & $0.892$ & $0.909$ & $0.918$ \\\hline\hline
		\multirow{2}{*}{\textbf{Train}} & \multicolumn{2}{c||}{\textbf{WE}} & \multicolumn{2}{c||}{\textbf{AF}} & \multicolumn{2}{c}{\textbf{QF}} \\\cline{2-7}
			 & \textbf{$\textbf{SS}^*$} & \textbf{$\textbf{BOW}^*$} & \textbf{$\textbf{SS}^*$} & \textbf{$\textbf{BOW}^*$} & \textbf{$\textbf{SS}^*$} & \textbf{$\textbf{BOW}^*$}\\\hline
			
			\textbf{20} & $\boldsymbol{0.754}$ & $0.709$ & $\boldsymbol{0.746}$ & $0.686$ & $\boldsymbol{0.716}$ & $0.645$\\\hline
			\textbf{50} & $\boldsymbol{0.77}$ & $0.774$ & $\boldsymbol{0.798}$ & $0.753$ & $\boldsymbol{0.764}$ & $0.702$\\\hline
			\textbf{100} & $\boldsymbol{0.83}$ & $0.807$ & $\boldsymbol{0.851}$ & $0.848$ & $\boldsymbol{0.812}$ & $0.783$\\\hline
			\textbf{200} & $0.863$ & $0.865$ & $\boldsymbol{0.907}$ & $0.876$ & $\boldsymbol{0.856}$ & $0.85$\\\hline
			\textbf{500} & $0.93$ & $0.921$ & $\boldsymbol{0.936}$ & $0.928$ & $\boldsymbol{0.9}$ & $0.9$\\\hline
			\textbf{1000} & $0.935$ & $0.947$ & $0.942$ & $0.949$ & $0.929$ & $0.934$\\\hline		
		\end{tabular}
	\end{center}
\end{table}	

\textbf{Does the choice of a word clustering algorithm make a difference}? In Table \ref{tab:bcvsw2v} we compare the best accuracy obtained by \textbf{BC} clustering to the best accuracy by \textbf{W2V} clustering. For both \textbf{BC} and \textbf{W2V}, we only considered $\textbf{D}_{50K}$, $\textbf{D}_{50Kp}$, and $\textbf{D}_{3K}$ corpuses. None of the algorithms is superior over all 6 disasters. \textbf{W2V} is superior on \textbf{OT}, \textbf{AF}, and \textbf{QF}, \textbf{BC} is superior on \textbf{SH}, and they are similar on the remaining two disasters. This is a slightly surprising result, considering the age of Brown clustering and the recent hype surrounding the word2vec algorithm. Studying this result more in depth would certainly be worthwhile. For the purposes of this paper, our conclusion is that both algorithms have their strengths, but that if we have to choose one, it would be \textbf{W2V}. 

\begin{table}[!htbp]
	\begin{center}
	\small
	\addtolength{\tabcolsep}{-3pt}
	\caption{Comparing two clustering algorithms \textbf{BC} and \textbf{W2V}.}
	\label{tab:bcvsw2v}
		\begin{tabular}{l|c|c||c|c||c|c}
		\hline
		\hline
		\multirow{2}{*}{\textbf{Train}} & \multicolumn{2}{c||}{\textbf{SH}} & \multicolumn{2}{c||}{\textbf{BB}} & \multicolumn{2}{c}{\textbf{OT}}  \\\cline{2-7}
			 & \textbf{$\textbf{BC}$} & \textbf{$\textbf{W2V}$} & \textbf{$\textbf{BC}$} & \textbf{$\textbf{W2V}$} & \textbf{$\textbf{BC}$} & \textbf{$\textbf{W2V}$} \\\hline
			
			\textbf{20} & $\boldsymbol{0.734}$ & $0.65$ & $0.707$ & $\boldsymbol{0.714}$ & $0.639$ & $\boldsymbol{0.682}$ \\\hline
			\textbf{50} & $\boldsymbol{0.758}$ & $0.736$ & $\boldsymbol{0.761}$ & $0.755$ & $0.706$ & $\boldsymbol{0.741}$ \\\hline
			\textbf{100} & $\boldsymbol{0.792}$ & $0.764$ & $\boldsymbol{0.794}$ & $0.784$ & $0.762$ & $\boldsymbol{0.789}$ \\\hline\hline
		\multirow{2}{*}{\textbf{Train}} & \multicolumn{2}{c||}{\textbf{WE}} & \multicolumn{2}{c||}{\textbf{AF}} & \multicolumn{2}{c}{\textbf{QF}} \\\cline{2-7}
			 & \textbf{$\textbf{BC}$} & \textbf{$\textbf{W2V}$} & \textbf{$\textbf{BC}$} & \textbf{$\textbf{W2V}$} & \textbf{$\textbf{BC}$} & \textbf{$\textbf{W2V}$}\\\hline
			
			\textbf{20} &  $0.73$ & $\boldsymbol{0.754}$ & $0.575$ & $\boldsymbol{0.613}$ & $0.591$ & $\boldsymbol{0.651}$\\\hline
			\textbf{50} & $\boldsymbol{0.759}$ & $0.748$ & $0.617$ & $\boldsymbol{0.678}$ & $0.668$ & $\boldsymbol{0.702}$\\\hline
			\textbf{100}& $0.801$ & $\boldsymbol{0.82}$ & $0.681$ & $\boldsymbol{0.689}$ & $0.724$ & $\boldsymbol{0.761}$\\\hline
		\end{tabular}
	\end{center}
\end{table}	

\textbf{What is the impact of the unlabeled corpus choice}? In Table \ref{tab:corpuses} we are showing the accuracies of the best \textbf{W2V} algorithm for four different choices of an unlabeled corpus: $\textbf{D}_{50K}$, $\textbf{D}_{50Kp}$, $\textbf{D}_{3K}$, and $\textbf{D}_{400M}$. On four disasters (\textbf{SH}, \textbf{OT}, \textbf{AF}, \textbf{QF}), the 400 million Twitter corpus results in the best accuracy, it is competitive on \textbf{WE}, and only on \textbf{BB} it is less accurate than $\textbf{D}_{50K}$. This result is a strong indicator that mantra ``the more data the better'' works in the emerging disaster scenario. Confirming this, it is interesting to see that using $\textbf{D}_{50K}$ is in general better than using its positively labeled subset $\textbf{D}_{50Kp}$. Finally, using the location-based tweets from the emerging disaster itself does not seem to be particularly effective, and it could probably be attributed to the small size of this corpus. 

\begin{table}[!htbp]
	\begin{center}
	\small
	\addtolength{\tabcolsep}{-3pt}
	\caption{Comparing 4 types of unlabeled corpus.}
	\label{tab:corpuses}
		\begin{tabular}{l|c|c|c|c||c|c|c|c}
		\hline
		\hline
		\multirow{2}{*}{\textbf{Train}} & \multicolumn{4}{c||}{\textbf{SH}} & \multicolumn{4}{c}{\textbf{BB}} \\\cline{2-9}
			 & \textit{3K} & \textit{50Kp} & \textit{50K} & \textit{400M} & \textit{3K} & \textit{50Kp} & \textit{50K} & \textit{400M}\\\hline
	\textbf{20} & $0.646$ & $0.693$ & $0.665$ & $\boldsymbol{0.696}$ & $0.694$ & $0.658$ & $\boldsymbol{0.714}$ & $0.662$  \\\hline
			\textbf{50} & $0.681$ & $0.726$ & $0.736$ & $\boldsymbol{0.746}$ & $0.722$ & $0.683$ & $\boldsymbol{0.755}$ & $0.737$\\\hline
			\textbf{100} & $0.661$ & $0.776$ & $0.764$ & $\boldsymbol{0.795}$ & $0.737$ & $0.733$ & $\boldsymbol{0.784}$ & $0.768$\\\hline\hline
		\multirow{2}{*}{\textbf{Train}} & \multicolumn{4}{c||}{\textbf{OT}} & \multicolumn{4}{c}{\textbf{WE}}  \\\cline{2-9}
			 & \textit{3K} & \textit{50Kp} & \textit{50K} & \textit{400M} & \textit{3K} & \textit{50Kp} & \textit{50K} & \textit{400M}\\\hline
	\textbf{20}  & $0.604$ & $0.651$ & $0.682$ & $\boldsymbol{0.688}$ & $0.661$ & $0.702$ & $\boldsymbol{0.754}$ & $0.71$  \\\hline
			\textbf{50} & $0.692$ & $0.708$ & $0.741$ & $\boldsymbol{0.746}$ & $0.719$ & $0.702$ & $0.748$ & $\boldsymbol{0.752}$ \\\hline
			\textbf{100}& $0.695$ & $0.732$ & $0.789$ & $\boldsymbol{0.795}$ & $0.774$ & $0.777$ & $\boldsymbol{0.82}$ & $\boldsymbol{0.82}$  \\\hline\hline
		\multirow{2}{*}{\textbf{Train}} & \multicolumn{4}{c||}{\textbf{AF}} & \multicolumn{4}{c}{\textbf{QF}} \\\cline{2-9}
			 &  \textit{3K} & \textit{50Kp} & \textit{50K} & \textit{400M} & \textit{3K} & \textit{50Kp} & \textit{50K} & \textit{400M}\\\hline
			\textbf{20} & $0.725$ & $0.608$ & $0.613$ & $\boldsymbol{0.746}$ & $0.608$ & $0.589$ & $0.651$ & $\boldsymbol{0.716}$ \\\hline
			\textbf{50} & $0.781$ & $0.636$ & $0.678$ & $\boldsymbol{0.798}$ & $0.636$ & $0.653$ & $0.702$ & $\boldsymbol{0.764}$\\\hline
			\textbf{100} & $0.811$ & $0.669$ & $0.689$ & $\boldsymbol{0.851}$ & $0.703$ & $0.698$ & $0.761$ & $\boldsymbol{0.812}$\\\hline
		\end{tabular}
	\end{center}
\end{table}	

\textbf{How many word clusters}? In Table 5 we show the best choice of the number of clusters for \textbf{W2V} obtained from the 400 million corpus. When the training data size is small, the smaller number of clusters is preferable. This result illustrates that the benefit of merging similar words into a smaller number of clusters to create dense features outweighs a potential loss of semantic cohesion in those clusters. As expected, for larger training sets, cluster cohesion becomes an important driving factor for high accuracy. 

\begin{table}[!htbp]
	\begin{center}
	\small
	\addtolength{\tabcolsep}{-2pt}
	\caption{The number of clusters (\textbf{K}) to achieve best performance for $\textbf{W2V}_{400M}$}
	\label{tab:kw2v400}
		\begin{tabular}{c|c|c|c|c|c|c}
		\hline
		\hline
		 \textbf{Train} & \textbf{SH} &\textbf{BB} & \textbf{OT} & \textbf{WE} & \textbf{AF} & \textbf{QF} \\\hline
		 \textbf{20}    & 500  & 2000 & 500  & 100  & 500  & 200 \\\hline
		 \textbf{50}    & 1000 & 1000 & 100  & 1000 & 1000 & 100 \\\hline
		 \textbf{100}   & 1000 & 2000 & 500  & 500  & 2000 & 1000\\\hline
		 \textbf{200}   & 2000 & 2000 & 500  & 2000 & 2000 & 2000\\\hline
		 \textbf{500}   & 2000 & 2000 & 2000 & 2000 & 1000 & 1000\\\hline
		 \textbf{1000}  & 2000 & 2000 & 2000 & 1000 & 2000 & 1000\\\hline
		\end{tabular}
	\end{center}
\end{table}

\textbf{Impact of training size}. Not to be lost in the previous discussion, it is evident that the number of labeled tweets has a strong influence on accuracy, and we can observe a robust increase in accuracy with the training data size.

\section{Conclusion}
In this paper we addressed the problem of retrieving disaster-related tweets shortly after the onset of a disaster. In such a scenario, we can expect to have access to a very limited number of labeled tweets. The accuracy of classifiers trained on such small data would be limited. To remedy this problem, we proposed a semi-supervised approach that can utilize a large unlabeled corpus of tweets to create word clusters and use them as features for classification. Our experiments on Twitter data from 6 disasters strongly indicate that the proposed semi-supervised approach could most often result in the improvements in accuracy as compared to the traditional supervised learning approach that uses feature selection on the bag of words features. Our study also provides useful insights into different modeling choices when using the proposed approach. While ``the bigger the better'' mantra in data science mostly holds true in this application, by a careful look at the results, it is also evident that ``one size does not fit all,'' and that many modelling choices have differing effect on different types of disasters. More effort is needed to gain a better insight into those differences and design approaches with more consistent behavior across a wide range of situational awareness scenarios. 

\section*{ACKNOWLEDGEMENTS}
This work was supported by NSF grant CNS-1461932.

\newpage
\bibliographystyle{abbrv}
\bibliography{sigproc}  
%
%

\end{document}